\documentclass[final]{cvpr}

\usepackage{times}
\usepackage{epsfig}
\usepackage{graphicx}
\usepackage{amsmath}
\usepackage{amssymb}
\usepackage{amstext}
\usepackage{subcaption}

\usepackage[pagebackref=true,breaklinks=true,colorlinks,bookmarks=false]{hyperref}

\def\cvprPaperID{17} 
\def\confYear{CVPR 2021}

\begin{document}

\title{An Improved Attention for Visual Question Answering}


\author{Tanzila Rahman$^{1,2}$ \qquad Shih-Han Chou$^{1,2}$ \qquad Leonid Sigal$^{1,2,3}$ \qquad Giuseppe Carenini$^{1}$\\
$^1$Department of Computer Science, University of British Columbia\\
Vancouver, BC, Canada\\
$^2$Vector Institute for AI \qquad
$^3$Canada CIFAR AI Chair \\
{\tt\small \{trahman8, shchou75, lsigal, carenini\}@cs.ubc.ca}
}

\maketitle

\begin{abstract}
   We consider the problem of Visual Question Answering (VQA). Given an image and a free-form, open-ended, question, expressed in natural language, the goal of VQA system is to provide accurate answer to this question with respect to the image. The task is challenging because it requires simultaneous and intricate understanding of both visual and textual information. Attention, which captures intra- and inter-modal dependencies, has emerged as perhaps the most widely used mechanism for addressing these challenges. In this paper, we propose an improved attention-based architecture to solve VQA. We incorporate an Attention on Attention (AoA) module within encoder-decoder framework, which is able to determine the relation between attention results and queries. Attention module generates weighted average for each query. On the other hand, AoA module first generates an \emph{information vector} and an \emph{attention gate} using attention results and current context; and then adds another attention to generate final \emph{attended information} by multiplying the two. We also propose multimodal fusion module to combine both visual and textual information. The goal of this fusion module is to dynamically decide how much information should be considered from each modality. Extensive experiments on VQA-v2 benchmark dataset show that our method achieves better performance than the baseline method.
\end{abstract}

\vspace{-0.2in}
\section{Introduction}
\begin{figure}[t!]
  \centering
  \includegraphics[scale=0.50]{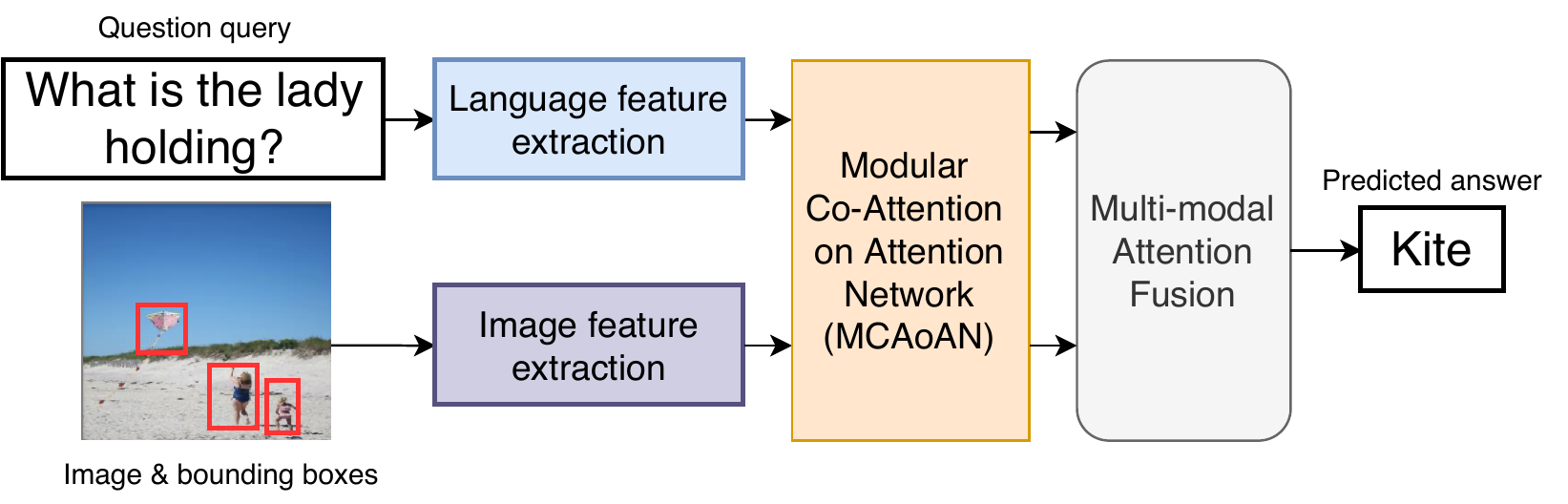}
\caption{\textbf{Illustration of our proposed framework.} Given an image and a query question, we first extract visual and language features respectively. Our proposed Modular Co-Attention on Attention Network (MCAoAN) takes the features as inputs and refines both features jointly. The multi-modal attention fusion fuses the refined visual and language features and then predicts final answer.}
\vspace{-0.25in}
\label{fig:intro}
\end{figure}

Different perceptual modalities can capture complementary information about aspects of an object, event or activity. As a result, multimodal representations are often shown to perform better in inference. 
Multimodal learning is widely used in the computer vision and forms basis for many visuo-lingual tasks, including image captioning~\cite{anderson2018bottom, rahman2019watch, xu2015show}, image-text matching~\cite{lee2018stacked, wang2018learning} and visual question answering~\cite{antol2015vqa, lu2016hierarchical}). 
Visual question answering (VQA) is perhaps the most challenging, 
requiring detailed and intricate image and textual understanding (see Figure~\ref{fig:intro}). Moreover, questions can be free-form and open-ended which requires VQA system to perform, simultaneously, a large collection of artificial intelligence tasks ({\em e.g.}, fine-grained recognition, object detection, activity recognition and visual common sense reasoning) to predict an accurate answer~\cite{antol2015vqa}. The answer format can also take different forms: a word, a phrase, yes/no, multiple choice, or a fill in the blank~\cite{srivastava2019visual}.

Inspired by the recent advantages of deep neural network, attention based approaches are widely used to solve many computer vision problems including VQA~\cite{anderson2018bottom, antol2015vqa, yu2018beyond}. An attention based approach for VQA was first introduced by Shih~\emph{et al.}~\cite{shih2016look} and nowadays it has become an essential component in most of the architectures. Recent works~\cite{lu2016hierarchical, yu2018beyond} include co-attention architecture to generate simultaneous attention in both visual and textual modality which increases prediction accuracy.
The limitation of these, more {\em global}, co-attention methods, is their inability to model interactions and attention among individual image regions and segments of text ({\em e.g.}, at the word token level). 

To address this problem, dense co-attention networks ({\em e.g.}, BAN~\cite{kim2018bilinear}, DCN~\cite{nguyen2018improved}) have been proposed, where each image region is able to interact with any (and all) words in the question. As a result, the models can get better understanding and reason about the image-question relationships; this, in turn, results in improved VQA performance. However, the bottleneck of these dense co-attention networks is the lack of self-attention within each modality, {\em e.g.}, region-to-region relationships in the image and word-to-word relationships in the question~\cite{yu2019deep}. 

To overcome this, Yu~\emph{et al.}~\cite{yu2019deep} proposed a deep Modular Co-Attention Network (MCAN) which consists of cascaded Modular Co-Attention (MCA) layers. MCA layer is obtained by combining two general attention units: self-attention (SA) and guided attention (GA). SA is able to capture intra-modal interactions ({\em e.g.}, region-to-region and word-to-word) while GA can capture cross-modal interactions ({\em e.g.}, word-to-region and region-to-word) by using multi-head attention architecture. 
While expressive and highly flexible, this form of attention still has a limitations. 
Specifically, the result is {\em always} a weighted combination of value pairs among which the model is attending. This maybe problematic when there is no closely related context over which the model is attending ({\em e.g.}, a word for which no context word or image region exists). In such a case attention would result in a noisy or, worse, distracting output vector that can negatively impact the performance.  


Motivated by Huang~\emph{et al.}~\cite{huang2019attention}, in this paper we leverage the idea of Attention on Attention (AoA) module to address the above mentioned limitation.
The AoA module is cascaded several times to form a novel Modular Co-Attention on Attention Network (MCAoAN) which is an improved extension to Modular Co-Attention Network (MCAN)~\cite{yu2019deep}.
The AoA module generates an information vector and an attention gate by using two separate linear transformations~\cite{huang2019attention} which is similar to GLU~\cite{dauphin2017language}. Attention results and query context are concatenated together and through a linear transformation we can obtain an information vector. Similarly through another linear transformation followed by a sigmoid activation function we can obtain an attention gate. By applying element-wise multiplication, we finally obtain attended information which builds relation between multiple attention heads and keep only the most related one discarding all irrelevant attention results. As a result, the model is able to predict more accurate answer. We also propose a multi-modal fusion mechanism to dynamically modulate modality importance while combining image and language features.

\vspace{0.1in}
\noindent
\textbf{Contributions.} Our contributions are:
\begin{itemize}
  \item We introduce an Attention on Attention module to form a Modular Co-attention on Attention Network (MCAoAN). MCAoAN captures intra- and inter-modal attention within and among visual and language modalities as well as able to mitigate information flow from irrelevant context. 
  
  
  \item We also present a multimodal attention-based fusion mechanism to incorporate both image and question features. Our fusion network dynamically decides how to weight each modality to generate final feature representation to predict the correct answer.
  
  \item Extensive experiments on the VQA-v2 benchmark dataset~\cite{goyal2017making} illustrate that the proposed method outperforms competitors, establishing significantly better performance than the baseline methods in visual question answering.
\end{itemize}

\section{Related Works}
In this section, we first briefly describe existing approaches for visual question answering and later review classical approaches to fuse image and question features.

\subsection{Visual Question Answering}
Antol~\emph{et al.}~\cite{antol2015vqa} first introduced the task of visual question answering (VQA), by combining computer vision with natural language processing, to mimic human understanding about a particular visual environment. The model used a CNN for feature extraction and an LSTM for language processing. The features were combined using element-wise multiplication in service of classifying the answers. 

Over the last few years, a large number of deep neural networks have been proposed to improve the performance on VQA. Moreover, attention-based approaches became widely used to solve various sequence learning tasks, including VQA. The goal of attention module is to identify the most relevant part of image or textual content. Yang~\emph{et al.}~\cite{yang2016stacked} introduced an attention network to support multi-step reasoning for the image question answering task. A combination of bottom-up and top-down attention mechanism was presented in ~\cite{anderson2018bottom}. A set of salient image regions were proposed by bottom-up attention mechanism using Faster R-CNN~\cite{ren2015faster}. On the other hand, task specific context was used to predict an attention distribution by top-down mechanism over the image regions. A model-agnostic framework is proposed by Shah~\emph{et al.} ~\cite{shah2019cycle} which relies on cycle consistency to learn VQA model. Their model not only answers the posed question, but also generates diverse and semantically similar variations of questions conditioned on the answer. They enforce network to match the predicted answer with the ground truth answer to the original question. Wu~\emph{et al.}~\cite{wu2019differential} propose a differential networks (DN), a novel plug and play module where differences between pair-wise features are used to reduce noise and learn inter-dependency between features. To extract image and text feature, Faster R-CNN~\cite{ren2015faster} and GRU~\cite{chung2014empirical} are used respectively. Both features are refined by a differential module and finally combined to predict the answers.

Recently, co-attention based approaches are becoming popular. The goal of co-attention model is to learn image and question attention simultaneously. Lu~\emph{et al.}~\cite{lu2016hierarchical} introduced a co-attention network that jointly reasons about image and question attention in a hierarchical fashion. Yu~\emph{et al.}~\cite{yu2018beyond} proposed an architecture to reduce irrelevant features by applying self attention for question embedding and question conditioned attention for image embedding. Multi-modal attention is proposed in~\cite{lu2017co, schwartz2017high} to focus on images, questions or answers feature simultaneously. Recently, bilinear attention is proposed in~\cite{fukui2016multimodal, kim2016hadamard, yu2017multi} to locate more accurate objects. A multi-step dual attention for multimodal reasoning and matching is presented in~\cite{nam2017dual}. One major limitation of these co-attention based approaches is lack of dense interactions between different modalities. To overcome this limitation, dense co-attention based methods are proposed in~\cite{yu2019deep, kim2018bilinear}. But dense co-attention can generate irrelevant vector in scenarios where nothing is related to the query. To overcome the problem, motivated by~\cite{huang2019attention}, in this paper we combine Attention-on-Attention (AoA) module with Modular co-attention network to improve existing architecture. Our revised attention mechanism delivers significantly better performance in VQA.

\subsection{Fusion Strategies for VQA}
To combine multi-modal features, sophisticated fusion technique is required. 
Depending on the type of fusion, existing VQA models can be divided into two categories: linear and bilinear~\cite{wu2019differential}. Linear models use simple fusion approaches to combine image and question features. Simple element-wise summation and element-wise multiplication are used in~\cite{lu2016hierarchical, yang2016stacked} and ~\cite{li2016visual, nam2017dual} respectively. On the other hand, bilinear model uses  more fine-grained approache to fuse image and question features. Fukui~\emph{et al.}~\cite{fukui2016multimodal} used outer product to fuse multi-modal features. A low-rank projection followed by an element-wise multiplication is used by Kim~\emph{et al.}~\cite{kim2016hadamard}. A Multi-modal Factorized Bilinear (MFB) pooling approach with co-attention learning is proposed in~\cite{yu2017multi}. Wu~\emph{et al.}~\cite{wu2019differential} proposed a Differential Networks (DN) based Fusion (DF) approach which first calculates differences between image and textual feature elements and then combines the differential representations to predict final answer.

In this paper, we propose an attention-based multi-modal fusion to combine image and question features by dynamically deciding how much weight to put on each modality; the weighted features are used to predict final answer. 

\section{Our Approach}
Motivated by~\cite{huang2019attention}, in this paper we present Modular Co-Attention on Attention Network (MCAoAN) module which is an extension of Modular Co-Attention Network (MCAN)~\cite{yu2019deep}. 
MCAoAN consists of Modular Co-Attention on Attention (MCAoA) layer which is a composition of two primary attention units: Self Attention on Attention (SAoA) and Guided Attention on Attention (GAoA) unit. In this section, we first discuss SAoA and GAoA units in Section~\ref{subsec:subsection1} followed by Modular Co-Attention on Attention (MCAoA) layer in Section~\ref{subsec:subsection2}. Lastly we present our MCAoAN with multimodal fusion mechanism in Section~\ref{subsec:subsection3} and Section~\ref{subsec:subsection4} respectively.

\begin{figure}[h!]
\begin{subfigure}{.50\textwidth}
  \centering
  \includegraphics[scale=.57]{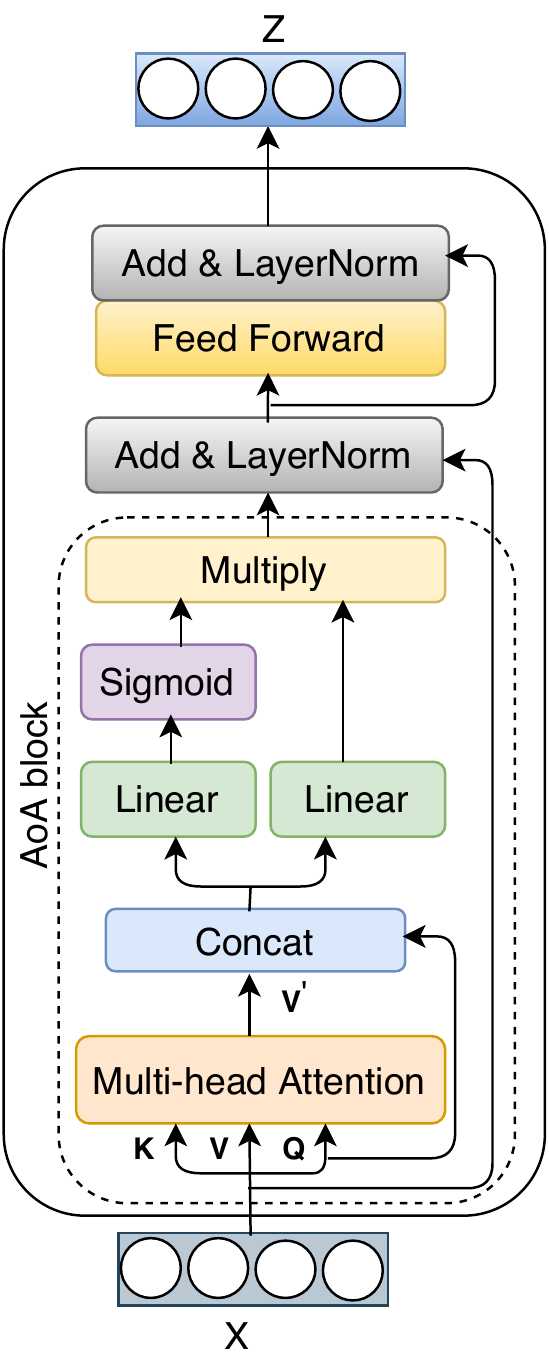}
  \caption{Self Attention on Attention block}
\end{subfigure}
\begin{subfigure}{.50\textwidth}
  \centering
  \includegraphics[scale=0.57]{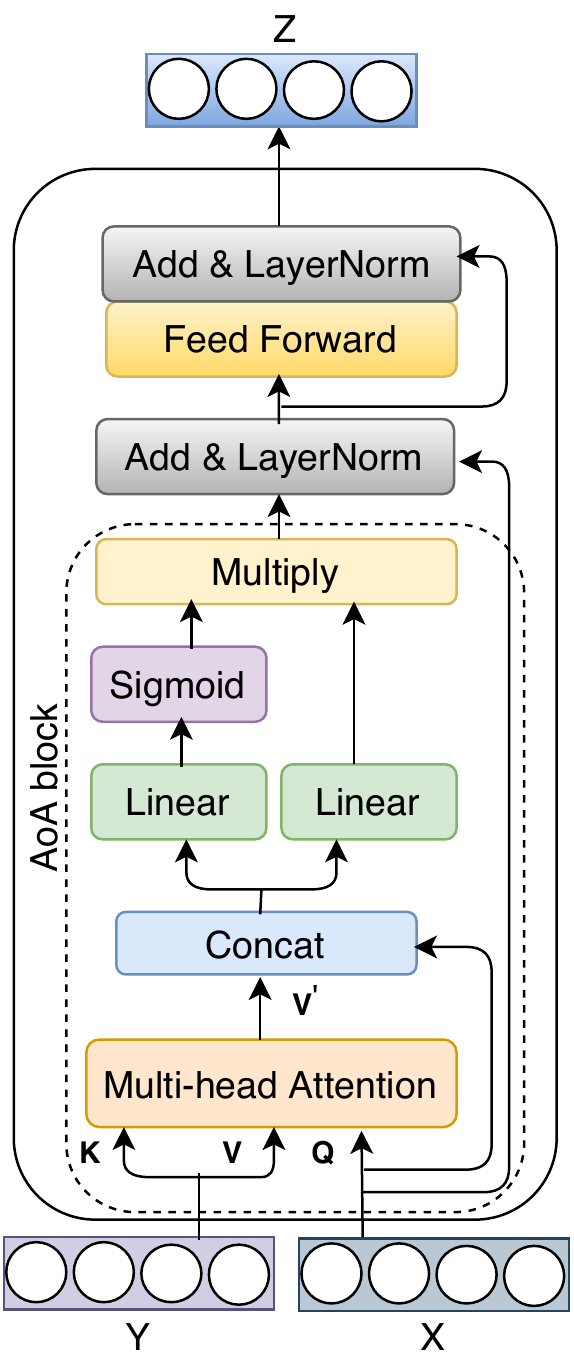}
  \caption{Guided Attention on Attention block}
\end{subfigure}
\vspace{-0.07in}
\caption{\textbf{Illustration of the two basic attention units.} (a) Self Attention on Attention block (SAoA), which takes input feature X and output attended feature Z for X; and (b) Guided Attention on Attention block (GAoA),which takes  two input features X and Y and generate attended feature Z for the input X guided by Y feature. Here X and Y represents image and question features respectively.}
\vspace{-0.07in}
\label{fig:attention_unit}
\end{figure}

\subsection{SAoA and GAoA Units}
\label{subsec:subsection1}
Our SAoA unit (see Figure~\ref{fig:attention_unit}(a)) is an extension of multi-head self attention mechanism~\cite{yu2019deep}. Multi-head attention consists of $N$ parallel heads where each head can be represented as a scaled dot product attention function as follows:
\vspace{-0.05in}
\begin{equation} \label{eq:mhattn}
    {\operatorname{f}}_{\operatorname{att}} = {\operatorname{f}}(Q,K,V) = {\text{\tt Softmax}}\left({\frac{QK}{\sqrt{d}}}\right)V,
\end{equation}
where attention function $\operatorname{f}(Q,K,V)$ operates on Q, K and V corresponds to query, key and value respectively. The output of this attention function is the weighted average vector $V^{\prime}$. To do so, first we calculate the similarity scores between Q and K; and normalize the scores with {\tt Softmax}. The normalized scores are then used together with V to generate weighted average vector $V^{\prime}$. Here, $d$ is the dimension of $Q$ and $K$; both dimensions are the same. 

The multi-head attention module always generates weighted vector, no matter whether it finds any relation between Q and K/V or not. So this approach can easily mislead or generate wrong answer for VQA. Therefore, following ~\cite{huang2019attention}, we incorporate another attention function over the multi-head attention module to measure the relation between attention results ($V^\prime$) and the query($Q$). The final AoA block will generate an information vector ($I$) and attention gate ($G$) through two separate linear transformations which can be represented as follows:

\begin{equation} \label{eq:AoA}
    {\operatorname{I}} = {\operatorname{W}}_{\operatorname{Q}}{\operatorname{Q}} + {\operatorname{W}}_{\operatorname{V}^{\prime}}{\operatorname{V}}^{\prime} + {\operatorname{b}}_{\operatorname{I}},
\end{equation}

\begin{equation} \label{eq:AoA_2}
    {\operatorname{G}} = \sigma ({\operatorname{W}}_{\operatorname{G}}{\operatorname{Q}} + {\operatorname{W}}_{\operatorname{G}^{\prime}}{\operatorname{V}}^{\prime} + {\operatorname{b}}_{\operatorname{G}}),
\end{equation}

\vspace{0.07in}

\begin{figure}[t!]
  \centering
  \includegraphics[scale=0.55]{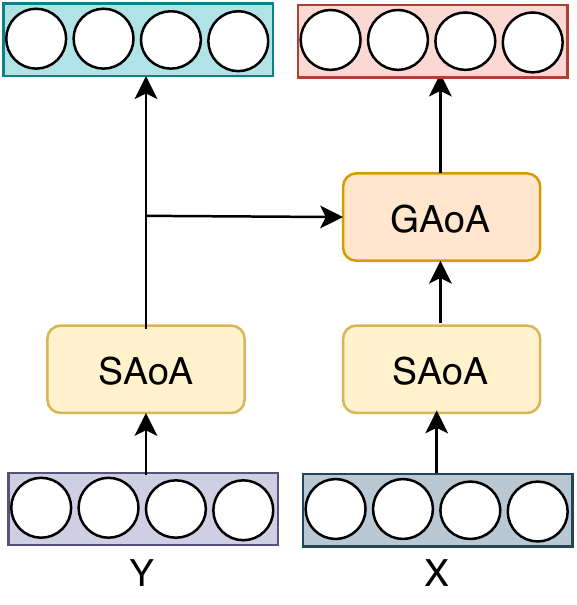}
  \vspace{-0.07in}
\caption{\textbf{Illustration of Modular Co-Attention on Attention (MCAoA) layer.} It consists of two attention units: Self Attention on Attention (SAoA) unit and Guided Attention on Attention (GAoA) unit where Y and X denotes question and image features respectively.}
\vspace{-0.07in}
\label{fig1:mca}
\end{figure}

Here, ${\operatorname{W}}_{\operatorname{Q}}$, ${\operatorname{W}}_{\operatorname{V}^{\prime}}$, ${\operatorname{W}}_{\operatorname{G}}$, ${\operatorname{W}}_{\operatorname{G}^{\prime}}$ $\in$ ${\mathbb{R}}^{d\times d}$ and ${\operatorname{b}}_{\operatorname{I}}$, ${\operatorname{b}}_{\operatorname{G}}$ $\in$ ${\mathbb{R}}^{d}$. $d$ is the dimension of $Q$ and $V^{\prime}$ where 
$V^{\prime}$ = ${\operatorname{f}}_{\operatorname{att}}$ and $\sigma$ denotes sigmoid function. AoA block adds another attention via element-wise multiplication between both information vector and attention gate. Moreover, SAoA uses a point-wise feed-forward layer after the AoA block, considering only input features $X = [x_1, x_2, ... , x_m] \in {\mathbb{R}}$.

Following ~\cite{yu2019deep}, we also propose another attention unit called guided attention on attention (GAoA) unit (see Figure~\ref{fig:attention_unit}(b)). Unlike SAoA unit, GAoA uses AoA block and a point-wise feed-forward layer along with two input features $X$ and $Y=[y_1, y_2, ... , y_n] \in {\mathbb{R}}$ where $X$ is guided by $Y$. In both attention unit, feed forward layer takes the output feature of AoA block and apply two fully connected layers along with ReLU and dropout function (i.e. $\text{\tt FC}(4d)-\text{\tt ReLU}-\text{\tt dropout}(0.1)-\text{\tt FC}(d)$). 

\subsection{MCAoA layers}
\label{subsec:subsection2}
Modular Co-Attention on Attention (MCAoA) layer (see Figure~\ref{fig1:mca}) consists of two attention units discussed in Section~\ref{subsec:subsection1}. Here $X$ and $Y$ represents image and question feature respectively. MCAoA layer is designed to handle multimodal interactions. We use cascaded MCAoA layers, {\em i.e.}, output from previous MCAoA is fed as input to the next MCAoA layer.  For both input features, MCAoA layer first uses two separate SAoA units to caption intra-modal interactions for $X$ and $Y$ separately and then uses GAoA unit to capture inter-modal relationships where $Y$ guides $X$ feature.     

\subsection{MCAoAN}
\label{subsec:subsection3}

\begin{figure*}[t!]
  \centering
  \includegraphics[scale=0.62]{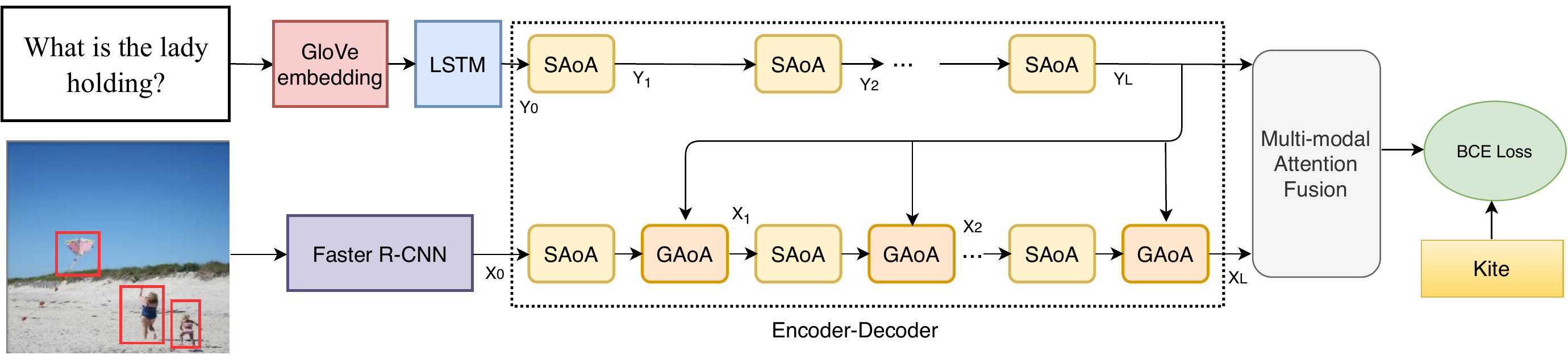}
\caption{\textbf{Illustration of overall architecture of proposed Modular Co-Attention on Attention Network (MCAoAN).} The network takes image and question feature as inputs. Image features are the intermediate features extracted from a Faster R-CNN~\cite{ren2015faster} model and each work from the question is transformed to a vector using 300-D GloVe word embedding~\cite{pennington2014glove} followed by a LSTM unit~\cite{hochreiter1997long}. Both features are fed to an Encoder-Decoder module consists of cascaded MCAoA layers and generate $X_L$ and $Y_L$ feature representations. $X_L$ and $Y_L$ denotes image and question feature respectively and combined together to generate desire answer by a multi-modal fusion module. }
\vspace{-0.07in}
\label{fig:main_architecture}
\end{figure*}

In this section, we discuss our proposed modular co-attention on attention network (MCAoAN) (see Figure~\ref{fig:main_architecture}) which is motivated by ~\cite{yu2019deep}. First we have to pre-process the inputs ({\em i.e.}, image and query question) into appropriate feature representations. We use Faster R-CNN~\cite{ren2015faster} with ResNet-101 as its backbone which is pretrained on Visual Genome dataset~\cite{krishna2017visual} to process input images. The intermediate feature of the detected object from Faster R-CNN is considered as visual feature representation. Following ~\cite{yu2019deep}, we also consider a threshold value to obtain dynamic number of detected objects, {\em e.g.}, $x_i$ is corresponds to i-th object feature. The final image feature is represented by a feature matrix $X$. 

The input query question is first tokenized and later trimmed to maximum 14 words. The pre-trained GloVe embedding~\cite{pennington2014glove} is used to transformed each word into a vector representation. This results a final representation of size $n\times300$ for a sequence of words where $n\in [1,14]$ denotes the number of word in the sequence. The word embedding is fed to a one layer LSTM network~\cite{hochreiter1997long} and generate final query feature matrix $Y$ by capturing the output features of all words.

Both input features are passed to the encoder-decoder module which contain cascaded MCAoA layers. Similar to ~\cite{yu2019deep}, encoder learns attention question features $Y_L$ by stacking $L$ number of SAoA units. On the other hand, decoder learns attended image features $X_L$ by stacking $L$ number of GAoA units by using query features $Y_L$. 
\newline

\subsection{Multi-modal Fusion. }
\label{subsec:subsection4}
\begin{figure*}[t]
  \centering
  \includegraphics[scale=0.47]{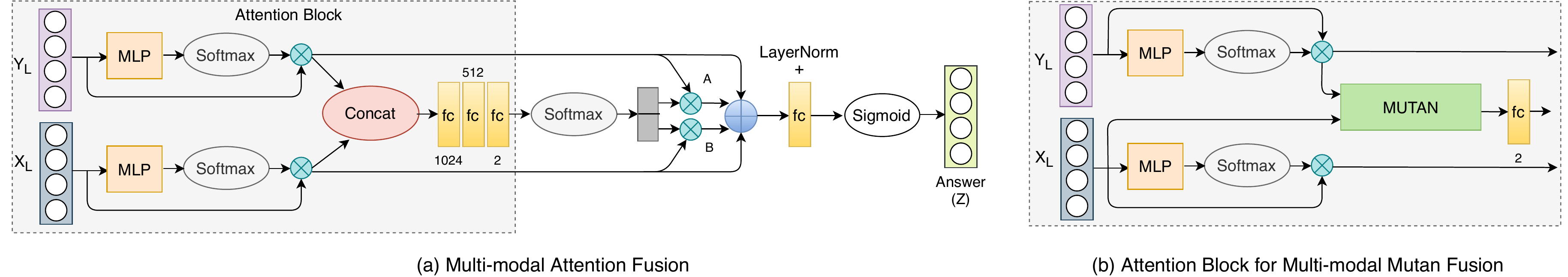}
\caption{\textbf{Illustration of proposed multimodal fusion network.} (a) Multi-modal attention fusion where we apply simple concatenation to combine initial attended features from both image and language modalities and apply series of fully connected layer to generate weighted features. The final weighted features represents how much importance should we give on each modality. (b) Multi-modal mutan fusion, another version of multi-modal fusion where we incorporate mutan fusion instead of concatenation keeping rest of the network similar to multi-modal attention fusion.}
\label{fig:fusion_network}
\end{figure*}

The outputs (i.e image features $X_L = [x_1, x_2, ..., x_m] \in {\mathbb{R}}^{m\times d}$ and question features $Y_L = [y_1, y_2, ...., y_n] \in {\mathbb{R}}^{n\times d}$) from encoder-decoder contains attended information about image and query regions. Therefore, we need to apply an appropriate fusion mechanism to combine both feature representation. In this paper, we propose two kind of multi-modal fusion networks (see Figure~\ref{fig:fusion_network}) to aggregate features of both modality: (1) Multi-modal Attention Fusion and (2) Multi-modal Mutan Fusion. Following~\cite{yu2019deep}, we first use two layers of MLP (i.e. $\text{FC(d)- ReLU - Dropout(0.1) - FC(1)}$) for both $X_L$ and $Y_L$; and generate attended features $X^\prime$ and $Y^\prime$ as follows:

\begin{equation} \label{eq:attx}
    \operatorname{X^\prime} = \sum_{i=1}^{m} \text{\tt Softmax}(\text{\tt MLP}(\operatorname{X_L}))\operatorname{x}_i,
\end{equation}
and
\begin{equation} \label{eq:atty}
    \operatorname{Y^\prime} = \sum_{i=1}^{n} \text{\tt Softmax}(\text{\tt MLP}(\operatorname{Y_L}))\operatorname{y}_i,
\end{equation}

Now we have rich attended features from both modality and at the same time each modality should use to generate attention with one another for better prediction. Therefore, we have to decide, how much information should use from each modality. Following ~\cite{mees2016choosing}, in multi-modal attention fusion, we apply concatenation on $\operatorname{X^\prime}$ and $\operatorname{Y^\prime}$ followed by a series of fully-connected layers ({\em i.e.}, $\text{\tt FC}(1024)-\text{\tt Dropout}(0.2)-\text{\tt FC}(512)-\text{\tt Dropout}(0.2)-\text{\tt FC}(2)-\text{\tt Softmax}$) (see Figure~\ref{fig:fusion_network} (a)). The output of these operations is a 2-dimensional feature vector that corresponds to the importance of two modality for answer prediction. After that, we generate weighted average of attended feature (i.e. $A$ and $B$) for each modality similar to eq.~\ref{eq:attx} and ~\ref{eq:atty}. $A$ and $B$ is combined with attended visual and textual features $X^\prime$ and $Y^\prime$. Finally, fused feature is fed to a $\text{\tt LayerNorm}$ to stabilize the training followed by a fully connected layer and sigmoid activation to generate predicted answer $Z$. We use binary cross-entropy loss (BCE) to train the network. 

On the other hand, we also leverage a powerful fusion technique, MUTAN fusion~\cite{ben2017mutan}, to integrate image and question features (see figure~\ref{fig:fusion_network} (b)) in multi-modal mutan fusion. The network is similar to the above model but replacing the concatenation to MUTAN fusion with fully-connected layers ({\em i.e.}, $\text{\tt Dropout}(0.2)-\text{\tt FC}(2)-\text{\tt Softmax}$).

\section{Experiments} 
In this section we first describe the dataset (see Section~\ref{sec:datasets}) used in our experiments. Then we present experimental setup and implementation details in Section~\ref{sec:experimental_setup}. In Section~\ref{sec:ablation_study}, we include a number of ablations to show the effectiveness of our proposed model. Lastly, we discuss experimental results in Section~\ref{sec:experimental_results}.

\begin{table}[t!]
\begin{center}
\begin{tabular}{lllll}
\hline
 \bf L & \bf All & \bf Other & \bf Y/N & \bf Num \\ \hline
 L = 2 & 81.88 & 74.47 & 96.11 & 69.00 \\
 L = 4 & 83.34 & \textbf{76.48} & 96.65 & 71.00\\
L = 6 & \textbf{83.45} & 76.45 & \textbf{96.83} & \textbf{71.44} \\
L = 8 & 82.20 & 75.42 & 95.87 & 68.53 \\
\hline
\end{tabular}
\end{center}
\vspace{-0.05in}
\caption{\label{tab:L-value} \textbf{Experimental results with different $L$.} Here we use a range of values from 2 to 8 on validation set. Best performance is achieved with $L=6$. Therefore, in this paper we choose  $L=6$ for our work.}
\vspace{-0.05in}
\end{table}

\begin{table*}[t!]
\begin{center}
\begin{tabular}{lllll}
\hline
 \bf Methods & \bf All & \bf Other & \bf Y/N & \bf Num \\ \hline
MCAN~\cite{yu2019deep} & 81.20 & 73.73 & 95.86 & 67.30 \\
Ours (MCAoAN) & 82.91 & 75.92 & 96.47 & 70.38 \\
Ours (MCAoAN + Mutan) & 83.00 & 76.13 & 96.36 & \textbf{70.42}\\
Ours (MCAoAN + Multi-modal Attention Fusion) & \textbf{83.25} & \textbf{76.51} & \textbf{96.58} & 70.40 \\
\hline
\end{tabular}
\end{center}
\vspace{-0.05in}
\caption{\label{tab:table1} \textbf{Visual Question Answering results using VQA-v2 dataset}. Comparison of our proposed approach with state-of-the-art method on validation set. Here we also show each component in our proposed method contribute to increase the performance of VQA system. }
\vspace{-0.07in}
\end{table*}
\begin{table}[t!]
\begin{center}
\begin{tabular}{lllll}
\hline
\bf Methods & \bf All & \bf Other & \bf Y/N & \bf Num \\ \hline
Bottom-up~\cite{teney2018tips} & 65.32 & 56.05 & 81.82 & 44.21 \\
MFH~\cite{yu2018beyond} & 68.76 & 59.89 & 84.27 & 49.56 \\
BAN~\cite{kim2018bilinear} & 69.52 & 60.26 & 85.31 & 50.93 \\
BAN+Counter~\cite{kim2018bilinear} & 70.04 & 60.52 & 85.42 & 54.04 \\
MuRel~\cite{cadene2019murel} & 68.03 & 57.85 & 84.77 & 49.84\\
MCAN~\cite{yu2019deep} & 70.63 & 60.72 & 86.82 & 53.26 \\
Ours (MCAoA) & \textbf{70.90} & \textbf{60.97} & \textbf{87.05} & \textbf{53.81} \\


\hline
\end{tabular}
\end{center}
\vspace{-0.05in}
\caption{\label{tab:table2} \textbf{Experimental results with other state-of-the-art models on Test-dev.} }
\vspace{-0.05in}
\end{table}

\begin{table}[t!]
\begin{center}
\begin{tabular}{lllll}
\hline
 \bf Methods & \bf All & \bf Other & \bf Y/N & \bf Num \\ \hline
Bottom-up~\cite{teney2018tips} & 65.67 & 56.26 & 82.20 & 43.90 \\
BAN+Counter~\cite{kim2018bilinear} & 70.35 & - & - & - \\
MuRel~\cite{cadene2019murel} & 68.41 & - & - & -\\
MCAN~\cite{yu2019deep} & 70.90 & - & - & - \\
Ours (MCAoA) & \textbf{71.14} & \textbf{61.18} & \textbf{87.25} & \textbf{53.36} \\
\hline
\end{tabular}
\end{center}
\vspace{-0.05in}
\caption{\label{tab:table3} \textbf{Experimental results with other state-of-the-art models on Test-std.} }
\vspace{-0.2in}
\end{table}

\subsection{Datasets}\label{sec:datasets}
To evaluate our method, in this paper we use VQA-v2 benchmark dataset~\cite{goyal2017making} which consists of images from MS-COCO dataset~\cite{lin2014microsoft} with human annotated question-answer pairs. There are 3 questions for each image and 10 answers per questions. The dataset has three parts: train set (80k images with 444k QA pairs), validation set (40k images with 214k QA pairs) and test set (80k images with 448k QA pairs). Moreover, test set is splited into two subsets: test-dev and test-standard where both are used for online evaluation performance. For measuring the overall accuracy, three types of answer are considered: Number, Yes/No and other.

\subsection{Experiment and Implementation Details}\label{sec:experimental_setup}
To evaluate our method, we follow the experimental protocol proposed by~\cite{yu2019deep}. The number of head in multi-head attention is 8. The latent dimension for both multi-head and AoA block is 512. Therefore, the dimension of each head is $512/8 = 64$. The size of the answer vocabulary is 3129. 

To train the MCAoA network we use Adam solver with $\beta_1 = 0.9$ and $\beta_2 = 0.98$. We train our network up to 13 epoch with batch size 64 which takes around 24hrs to complete the training. The learning rate set to $min(2.5Te^{-5} , 1e^{-4} )$ where $T$ represents current epoch. Learning rate starts to decay by $1/5$ every $2$ epochs when $10 \leq T$.

\subsection{Ablation studies}\label{sec:ablation_study} 
We run a number of experiments to show the effectiveness of our proposed method and results of these experiments are presented in Table~\ref{tab:L-value} and \ref{tab:table1}. 

\vspace{0.07in}
\noindent
\textbf{Number of Cascaded Layer ($L$):} MCAoA layers consist of $L$ number of stacked SAoA and GAoA units. From Table~\ref{tab:L-value}, we can see, initially, with the increasing value of $L$, performance of the model is also increasing -- up to $L=6$. After that the performance is saturated. We use $L=6$ in our final model. We use~\emph{validation set} for this experiment with the default hyperparameters of ~\cite{yu2019deep}.

\begin{figure*}[t!]
  \centering
  \includegraphics[scale=0.69]{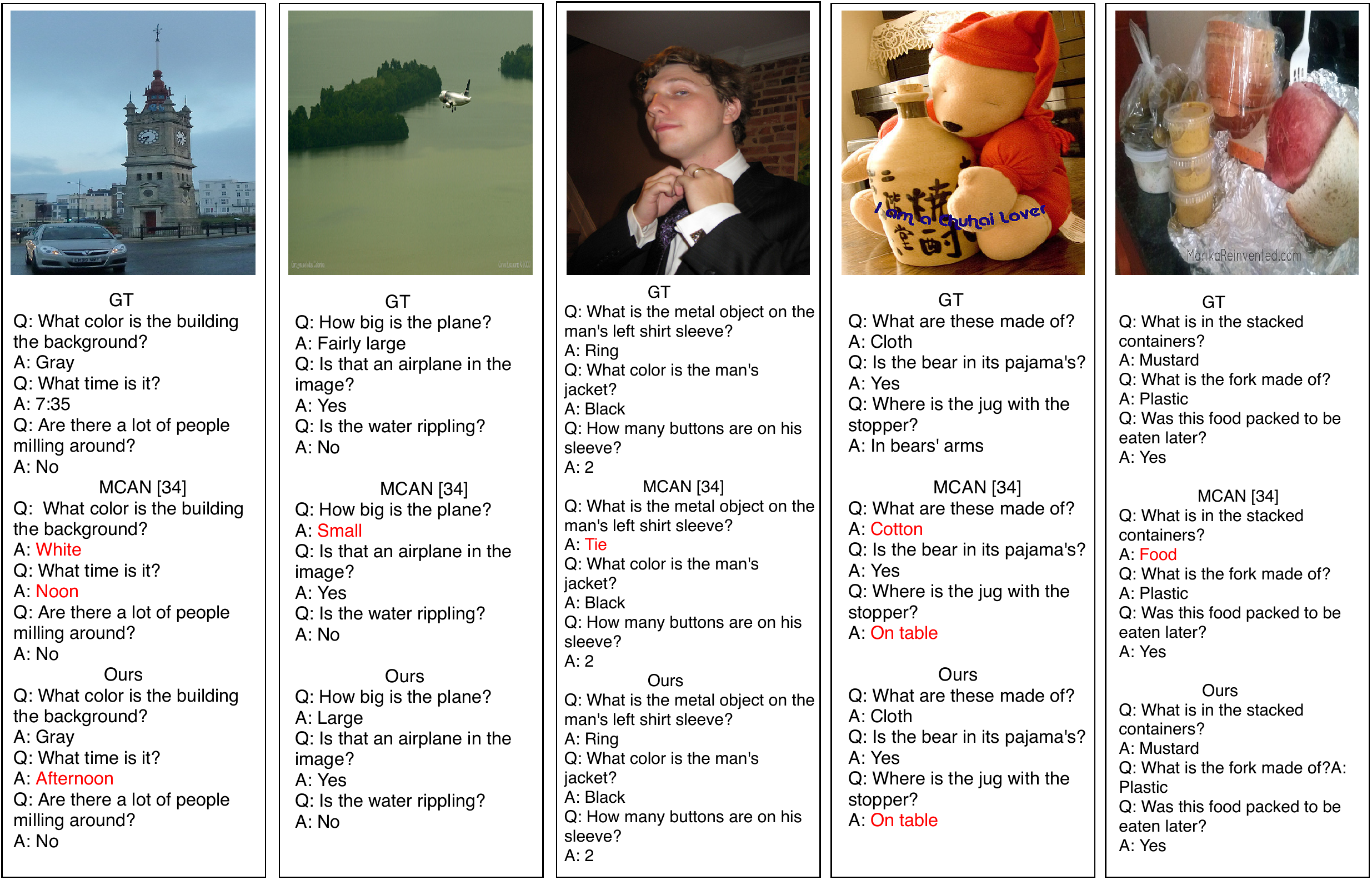}
\caption{\textbf{Illustration of some qualitative results from validation set using MCAN~\cite{yu2019deep} and our method.} First we present ground-truth (GT) annotations followed by the predicted answers of state-of-the-art method and our proposed method. Here Q and A represents query question and generated answer respectively. Moreover, red text indicates predicted wrong answer for the corresponding question.}
\label{fig:qualitative_results}
\end{figure*}

\begin{figure*}[h]
  \centering
  \includegraphics[scale=0.2]{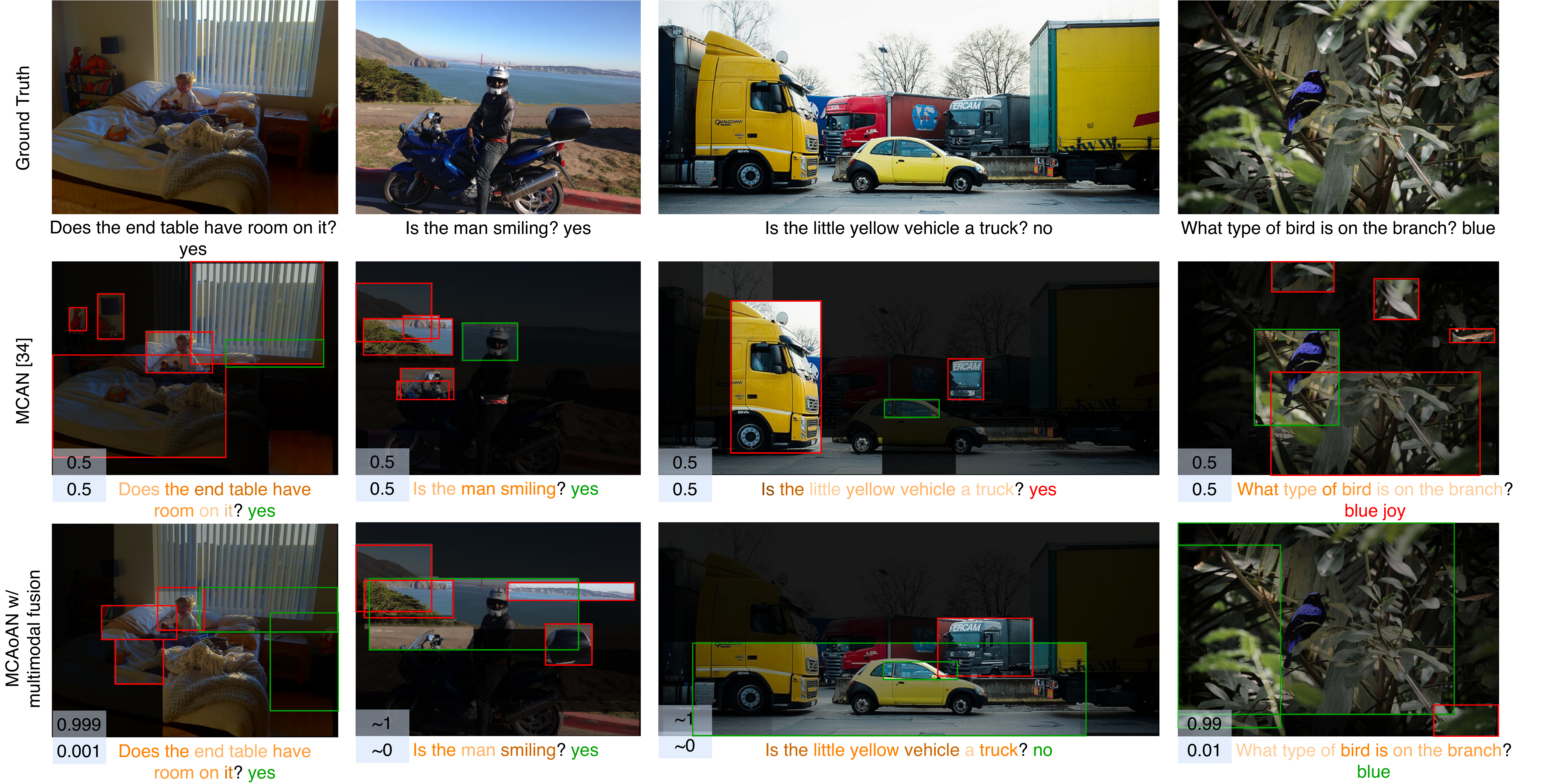}
\caption{\textbf{Qualitative results with multi-modal fusion.} The first row is the input images, questions and ground truth answers. The second row is the baseline model MCAN~\cite{yu2019deep}. The third row is the proposed model, MCAoAN w/ multi-modal fusion. The probabilities on the image and in front of the question represent the weight from each modality. We also show the attention across bounding boxes and words. In the image, the brighter area with green bbox has higher weight. For questions, the darker color of the word, the higher attention score.}
\label{fig:qualitative}
\end{figure*}

\begin{figure*}[h]
  \centering
  \includegraphics[scale=0.69]{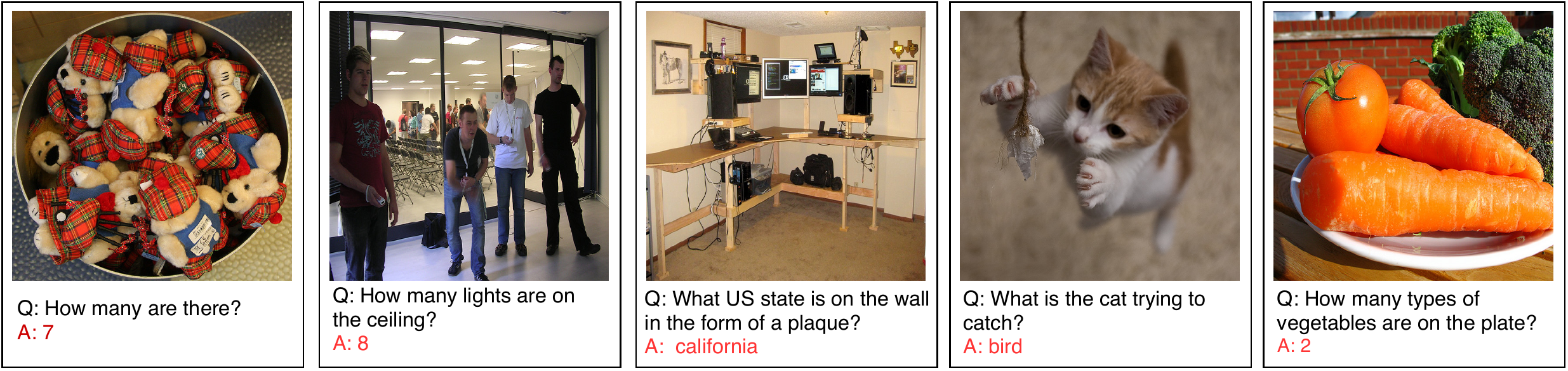}
\caption{\textbf{Illustration of some failure cases using our method.} Here Q and A represents query question and predicted wrong answer (mark as red) respectively.}
\label{fig:failure_cases}
\end{figure*}

\noindent
\textbf{Effectiveness of Each Individual Component: } In this paper, our improved architecture has two important components: (1) MCAoAN network which consists of SAoA module and GAoA module and (2) Multi-modal fusion to incorporate image and language features. Here, we describe two different fusion mechanism : Mutan fusion and Multi-modal attention fusion. Table~\ref{tab:table1} shows experimental results of these individual components and compare with existing MCAN~\cite{yu2019deep} on ~\emph{validation set}. From the table, we see that incorporating SAoA and GAoA module with MCAN improves the performance of VQA system. 

Moreover, we argue that a sophisticated way to aggregate language and visual features to support multi-modal reasoning is essential to further boost the performance. Table~\ref{tab:table1} also shows the comparison of different fusions with the MCAoA only where the former achieves better performance. More specifically, 
our proposed MCAoAN with both multi-modal fusion modules outperforms the baseline about $2\%$ accuracy on the whole~\emph{validation set}. This shows that the fusion module is important to combine vision and language representations. The proposed both fusion modules are suitable for VQA tasks. Among them multi-modal attention fusion performs the best. Beside that, Table~\ref{tab:table1} also shows that each individual component within our proposed method is important to increase the performance of VQA system. 

\subsection{Experimental Results}\label{sec:experimental_results}
We evaluate our model on VQA-v2 dataset and compare with other state-of-the-art methods. We re-run the PyTorch implementation provided by~\cite{yu2019deep}\footnote{\url{https://github.com/MILVLG/mcan-vqa}} and compare the results with our proposed method. Table~\ref{tab:table2} and~\ref{tab:table3} shows experimental results using test-dev and test-std respectively using online evaluation~\footnote{\url{https://evalai.cloudcv.org/web/challenges/challenge-page/163/overview}}. Offline evaluation only supports on validation split (see table~\ref{tab:table1}). Figure~\ref{fig:qualitative_results}, shows some qualitative results using our method on validation set.  From the experimental results,  we can see that our proposed method outperforms other baseline methods on VQA. In Figure~\ref{fig:qualitative}, we also visualize multi-modal fusion to compare how correctly MCAN~\cite{yu2019deep} and our proposed multi-modal attention fusion can able to focus on image and question elements. The brighter bounding-box along with green color within the image and darker color in question represents higher attention score. We can see that our proposed method is able to focus more on correct answer. Beside that, Figure~\ref{fig:failure_cases} shows typical failure cases using our method. 

\vspace{-0.10in}

\section{Conclusion}
\vspace{-0.07in}
In this paper, we propose an improved end-to-end attention based architecture for visual question answering. Our proposed method includes modular co-attention on attention module with multi-modal fusion architecture. In this paper, we propose two version of multi-modal fusion : multi-modal attention fusion and multi-modal mutan fusion. Experimental results show that each component within our model improve the performance of VQA system. Moreover, The final network achieves significant performance on VQA-v2 dataset.

{\small
\bibliographystyle{ieee_fullname}
\bibliography{egbib}
}

\end{document}